\documentclass[conference]{IEEEtran}
\IEEEoverridecommandlockouts
\usepackage{cite}
\usepackage{amsmath,amssymb,amsfonts}
\usepackage{algorithmic}
\usepackage{graphicx}
\usepackage{textcomp}
\usepackage{xcolor}
\usepackage{mathtools}

\usepackage{multirow}
\usepackage[breaklinks, colorlinks]{hyperref}
\usepackage{booktabs}
\usepackage{caption}
\usepackage{subcaption}
\usepackage{makecell}
\def\BibTeX{{\rm B\kern-.05em{\sc i\kern-.025em b}\kern-.08em
    T\kern-.1667em\lower.7ex\hbox{E}\kern-.125emX}}
    
\begin{document}

\title{Bridging the Gap Between Saliency Prediction \\ and Image Quality Assessment
}

\author{
    \IEEEauthorblockN{
        Kirillov Alexey\IEEEauthorrefmark{1}\IEEEauthorrefmark{2}\quad 
        Andrey Moskalenko\IEEEauthorrefmark{1}\IEEEauthorrefmark{3}\IEEEauthorrefmark{4}\quad 
        Dmitriy Vatolin\IEEEauthorrefmark{1}\IEEEauthorrefmark{4}
    }
    \IEEEauthorblockA{\IEEEauthorrefmark{1}Lomonosov Moscow State University \qquad
    \IEEEauthorrefmark{2}Yandex\\
    \IEEEauthorrefmark{3}AIRI, Moscow, Russia \qquad \IEEEauthorrefmark{4}MSU Institute for Artificial Intelligence \\
    \tt\small \{alexey.kirillov, andrey.moskalenko, dmitriy\}@graphics.cs.msu.ru}
}

\maketitle


\begin{abstract}
Over the past few years, deep neural models have made considerable advances in image quality assessment (IQA). However, the underlying reasons for their success remain unclear due to the complex nature of deep neural networks. IQA aims to describe how the human visual system (HVS) works and to create its efficient approximations. On the other hand, Saliency Prediction task aims to emulate HVS by determining areas of visual interest. Thus, we believe that saliency plays a crucial role in human perception. 

In this work, we conduct an empirical study that reveals the relation between IQA and Saliency Prediction tasks, demonstrating that the former incorporates knowledge of the latter. Moreover, we introduce a novel SACID dataset of saliency-aware compressed images and conduct a large-scale comparison of classic and neural-based IQA methods. 
Supplementary code and data will be available at \href{https://huggingface.co/datasets/alexkkir/SACID}{https://huggingface.co/datasets/alexkkir/SACID}.

\end{abstract}

\begin{IEEEkeywords}
Image Quality Assessment, Visual Saliency Prediction, Explainable AI
\end{IEEEkeywords}


\section{Introduction}

Image Quality Assessment (IQA) aims to measure image quality aligned with human visual perception. Improving IQA can greatly enhance user experience in tasks such as image compression, restoration, editing, and generation.

Despite achieving high correlations with human judgments, the internal workings of IQA models remain unclear. An important open question is whether IQA models implicitly adopt properties of human vision, particularly visual saliency—the human tendency to focus attention on certain image regions~[\citen{saliency_bio}, \citen{iqa_hvs}]. Meanwhile, Saliency Prediction (SP) has advanced significantly, delivering accurate predictions of human attention. 

\begin{figure}[ht]
    \centering
    \includegraphics[width=0.99\columnwidth]{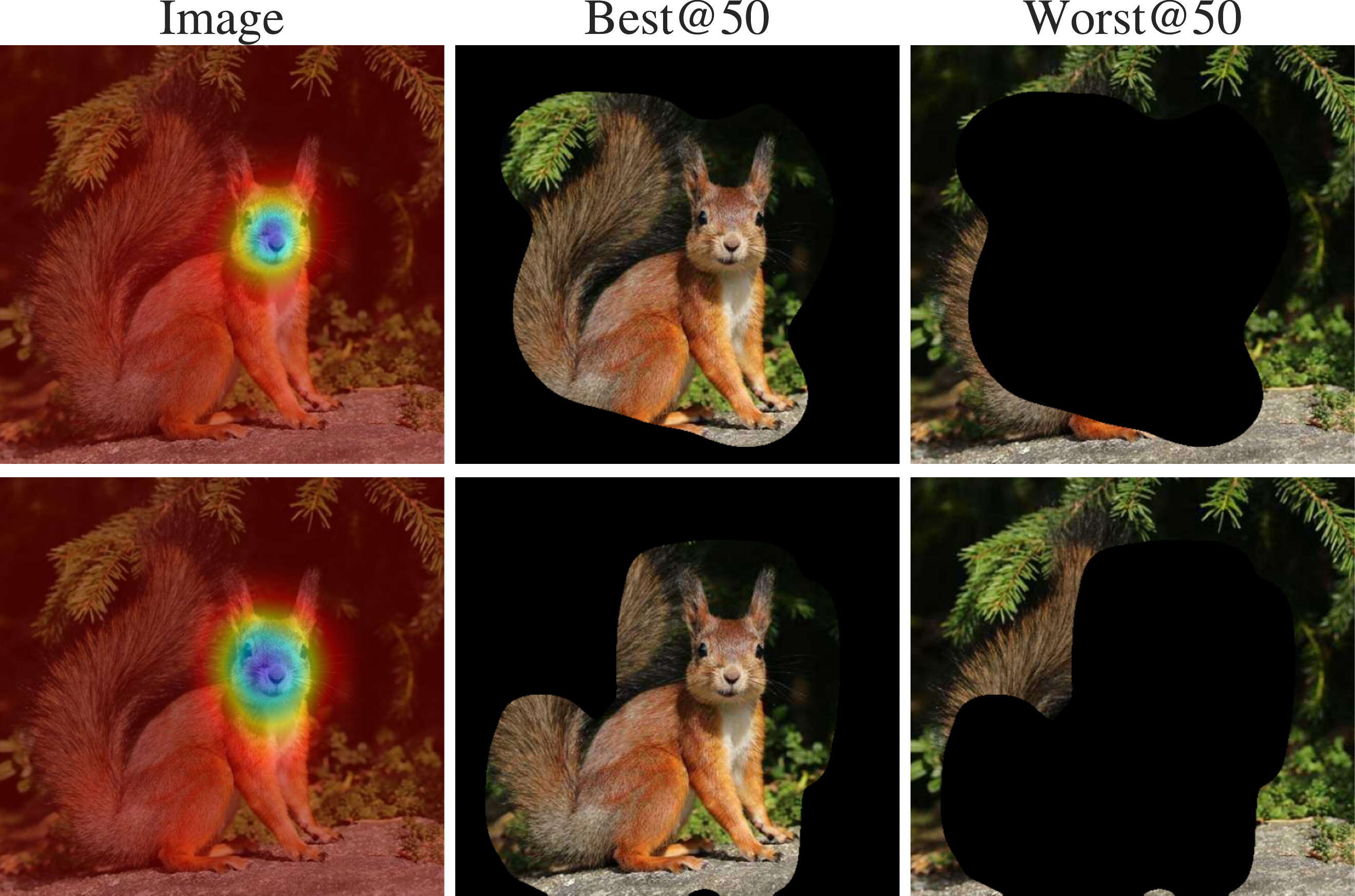}
    \caption{Saliency and GradCAM identify important regions. Upper row: saliency maps predicted by SOTA model TranSalNet~\cite{transalnet}. Bottom row: GradCAM extracted from our IQA model (Baseline-EfficientB0).}
    \label{fig:masked_images}
\end{figure}



In this work, we explore the connection of IQA and SP. Our main contributions are as follows:
\begin{itemize}
    \item We propose a methodology to extract saliency maps from trained IQA models, which reveals that learning-based IQA methods incorporate saliency in their predictions and can even outperform saliency prediction baselines, such as center-prior.
    \item We present a method for parameter-free dual-task training strategy for IQA and Saliency Prediction, which reveals that these tasks are connected and can be solved simultaneously without quality drops.
    \item We conduct a subjective study with 1400+ assessors to evaluate the effectiveness of existing IQA metrics for non-uniform compressed content employing various saliency-aware coding strategies.
\end{itemize}



\section{Related Work}


\subsection{Image Quality Assessment}

IQA aims to measure the perceptual quality of an image by assessing image artifacts and distortions. Most IQA techniques can be categorized as Full-Reference (FR) or No-Reference (NR), with the latter more relevant for real-world applications~\cite{survey_iqa}, where a reference is not available. 

Early approaches relied on natural scene statistics (NSS) [\citen{ssim}, \citen{ms_ssim}, \citen{vif}, \citen{brisque}, \citen{ew_ssim}] with hand-crafted features and regression models. As deep learning evolved, neural network-based approaches became popular, typically consisting of a convolutional or transformer backbone with a regression head. HyperIQA~\cite{hyperiqa} proposed a multiscale feature, while MUSIQ~\cite{musiq} employed multiscale inputs and a vision transformer backbone.

Some methods [\citen{topiq}, \citen{halfimage}] create paired models for both FR and NR scenarios, such as TOPIQ, which uses features from original, distorted, and difference images at each layer.


\subsection{Saliency in Image Quality Assessment}

Research indicates a connection between saliency and IQA [\citen{topiq}, \citen{wang_ugc_2023}, \citen{iqa_sal}, \citen{hantao_liu_visual_2011}, \citen{lin_saliency-aware_2023}]. Efforts to enhance IQA models by incorporating saliency include [\citen{iqa_sal}, \citen{sgdnet}, \citen{hvs5m}, \citen{cai_blind_2023}]. Early studies [\citen{ew_ssim}] improved NSS-based metrics by reweighting error maps based on saliency. SGDNet~\cite{sgdnet} introduced a separate head to predict saliency and reweight features.

TranSLA~\cite{sal_transformer} incorporates saliency as a query branch in cross-attention, while HVS-5M~\cite{hvs5m} uses a pre-trained SP model to reweight features. SCVC~\cite{scvs} aggregates patch scores using Gaussian functions and saliency maps (SM). LPIPS~\cite{lpips} performance can be enhanced by spatially reweighting feature maps~\cite{topiq}. However, some studies~\cite{wang_ugc_2023} report only slight improvement in VQA models from using saliency, and our research suggests models already incorporate saliency knowledge into their predictions.


\subsection{AI Explainability}

Neural networks, despite their high performance, remain a black box. Methods have been developed to provide insight into their internal operations. CAM~\cite{cam} considers a simple case of a CNN with a head consisting of a single linear layer after global average pooling (GAP). GradCAM~\cite{gradcam} extends CAM to models with arbitrary heads, using gradients w.r.t. the model's predictions. Subsequent works [\citen{gradcam_pp}, \citen{hirescam}, \citen{normgrad}] propose heuristics for improving the method.

Some researchers have studied benchmarking explanation maps. In~\cite{gradcam_pp}, they suggest masking important image areas and monitoring changes in model confidence. Work~\cite{sanity_checks_xai} introduced a metric, area over the perturbation curve, as a measure of explanation-map fidelity.

Several works address explainability in IQA. In~\cite{halfimage}, authors showed that IQA models only require half of an image to make accurate predictions. They divided each image into 12 square patches and examined predictions of the transformer-based IQA model as they masked various combinations. They found that using important regions preserves model quality, while using trivial regions decreases it. We simplify this approach and show that masking pixels based on saliency maps or GradCAM yields comparable results.


\section{Experiments}


\subsection{Extracting Saliency from IQA models}\label{saliency_extraction}

\begin{table}[tbp]
\tabcolsep=5pt
\centering
\caption{Performance comparison on SALICON\cite{salicon} dataset. GradCAM-extracted saliency maps from IQA models outperform center-prior baseline, but are surpassed by SOTA saliency prediction methods. The best results are \textbf{bold}, the second-best are \underline{underlined}, and the third best are \textit{italics}.
}
\begin{tabular}{clcccc}
\toprule
Type                     & Method             & NSS $\uparrow$ & SIM $\uparrow$       & CC $\uparrow$     & KLD $\downarrow$  \\ \midrule
\multirow{1}{*}{Dummy}                    
                    & Center Prior        & 0,582          & 0,534                & 0,541             & 0,784 \\ \cmidrule{1-6}
\multirow{5}{*}{\thead{GradCAM \\ from IQA}} & Baseline-EfficientNet & 0,604          & \textit{0,584}       & \textit{0,637}    & \textit{0,700} \\ 
                         & Baseline-ResNet50  & \textbf{0,621} & 0,555                & 0,590             & 0,782  \\ 
                         & TOPIQ~\cite{topiq}              & 0,597          & 0,552                & 0,572             & 0,759  \\ 
                         & DBCNN~\cite{dbcnn}              & 0,579          & 0,566                & 0,590             & 0,707  \\ 
                         & CLIP-IQA+~\cite{clipiqa}          & 0,584          & 0,550                & 0,589             & 0,741  \\ \cmidrule{1-6}
\multirow{2}{*}{SOTA SP}    & MSINet~\cite{msinet}             & \textit{0,612} & \underline{0,767} & \underline{0,891} & \textbf{0,252}  \\ 
                         & TranSalNet~\cite{transalnet}         & \underline{0,613}  & \textbf{0,776}        & \textbf{0,903}     & \underline{0,258}  \\ 
\bottomrule
\end{tabular}
\label{tab:gradcam_corr}
\end{table}

\begin{figure*}[tbp]
    \centering
    \includegraphics[width=0.99\textwidth]{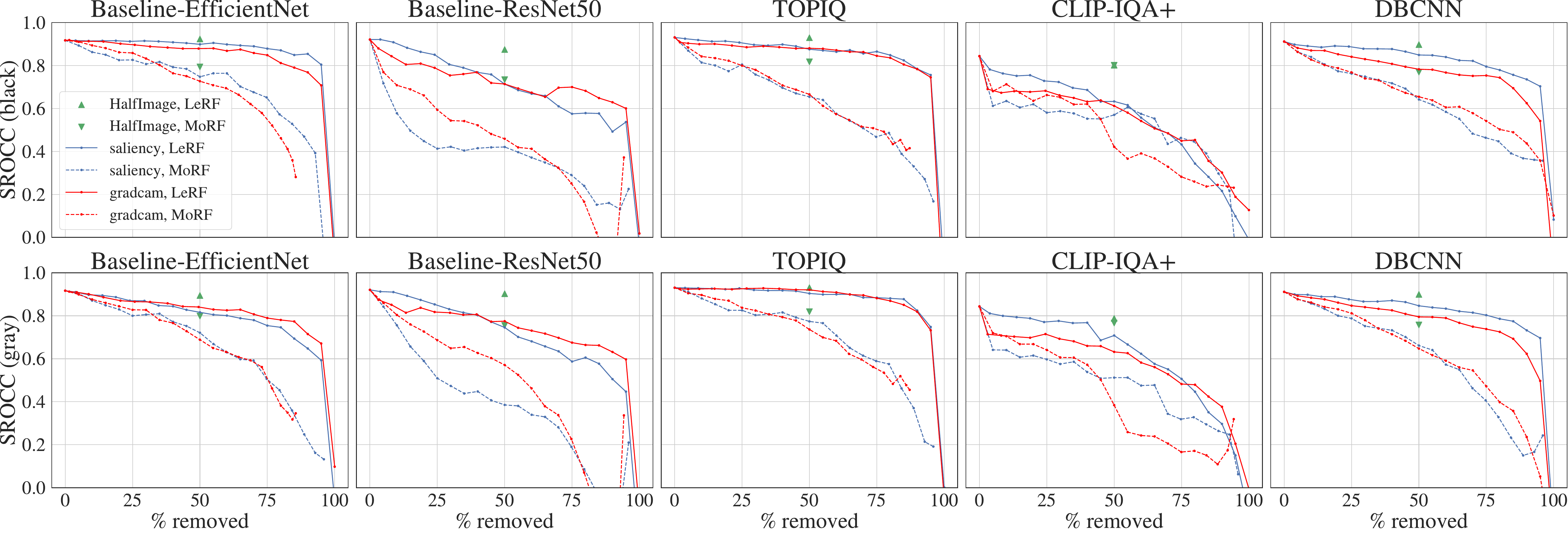}
    \caption{Masking input images according to saliency and GradCAM maps with black (upper row) and ImageNet mean (bottom row). We note a strong relation between the behavior of correlations when masking with saliency and GradCAM maps.}
    \label{fig:masking_plots}
\end{figure*}

We started our experiments by testing the hypothesis that ground-truth saliency (e.g. from eye-tracker) correlates with explanation GradCAM maps of IQA models, in other words, if it is possible to extract an approximation for saliency from a trained IQA model. In our experiments we used TOPIQ~\cite{topiq}, DBCNN~\cite{dbcnn} and CLIP-IQA~\cite{clipiqa}. We also trained our simple baseline model, consisting of a backbone, GAP pooling, and MLP head. We tested two backbones: EfficientNet-B0~\cite{efficientnet} and ResNet-50~\cite{resnet}. To build GradCAMs we used HiResCAM~\cite{hirescam} method, known to provably reflect the locations the model used for computation. Additionally, to improve the quality of maps and remove noise, we applied smoothing through augmentations and SVD decompositions of feature maps as recommended in~\cite{jacobgilpytorchcam}. Every image was passed six times through the model with small rotations. Then feature map from the last channel was channel-wise decomposed via SVD-decomposition and the first main component was taken.

We used straightforward saliency baselines -- Center Prior, a Gaussian distribution that approximates the saliency averaged over the dataset -- and two State-of-the-Art saliency prediction models [\citen{msinet}, \citen{transalnet}]. We evaluate the model quality based on common saliency metrics -- NSS, SIM, CC, and KLD calculated on the validation split of SALICON dataset~\cite{salicon}. Before calculating metrics, we employed map transforms~\cite{lyudvichenko2017semiautomatic} as a post-processing transformation.

Results are presented in Tab.~\ref{tab:gradcam_corr}. GradCAM better predicts saliency than center-prior, indicating that IQA models understand saliency distribution and allow extracting such proxy SM in a zero-shot mode.



\subsection{Saliency Masking}

Inspired by~\cite{halfimage}, we investigate the significance of individual image regions on image quality by masking them. The primary aim was to determine which interpretation maps best represent important areas of the image: saliency or GradCAM. For each image, the corresponding explanation map was used, with high values indicating important areas. Subsequently, all pixels in the image whose values did not exceed the threshold were masked.

Masking of different image portions was performed using two strategies: Most Relevant First (MoRF) and Least Relevant First (LeRF), as proposed in~\cite{sanity_checks_xai}. In MoRF, pixels with values above the threshold were masked, preserving trivial regions. Conversely, in LeRF, pixels with values below the threshold were masked, preserving important regions. Perturbations, such as filling with black color or ImageNet mean values, were applied. Before thresholding the explanation maps, Gaussian blur with a large kernel (approximately 101 pixels) and a small sigma was applied to ensure the calculation of thresholds corresponding to all quantiles. Fig.~\ref{fig:masked_images} shows examples of images whose regions are filled with black. For the experiment, images from the KonIQ-10k dataset were used. The performance of IQA models on masked images was calculated in terms of Spearman's Rank Order Correlation Coefficient (SROCC).

Fig.~\ref{fig:masking_plots} presents the results. A comparison was also made to the method from~\cite{halfimage}, referred to as "HalfImage".

Masking important areas (MoRF) decreases image quality more than masking trivial areas (LeRF). The same observation holds for GradCAM. Interestingly, masking with saliency results in a larger performance gap between LeRF and MoRF regions compared to HalfImage, even though the latter employs knowledge of pretrained IQA models. This finding suggests that saliency maps effectively highlight regions that are crucial for IQA models. Furthermore, it is noted that the behavior of models when masked with GradCAM maps and saliency exhibits significant similarity. This similarity can be interpreted as an existing relationship between the attention of IQA models and human visual attention.



\subsection{Dual-Task Training}

After discovering a close connection between IQA and saliency prediction (SP), we decided to train a model on both tasks simultaneously and determine whether we could do so without decreasing performance. We implemented two approaches to achieve this goal.

In the first approach, termed Baseline-Sal-Loss, we added a small decoder to our baseline model. It consisted of a single convolutional layer with a 1×1 kernel, followed by a sigmoid activation function to integrate saliency loss into the model.

The second approach, called Baseline-GradCAM-Loss, utilizes the GradCAM method as a secondary output, which we incorporated into the model's loss function to enhance saliency prediction. The GradCAM calculation is as follows:
\[
    L^{c}_{\text{GradCAM}} = \operatorname{ReLU} \left(\sum_k \alpha^c_k A^k \right),
    \quad
    \alpha^c_k = \frac{1}{Z} \sum_i \sum_j \frac{\partial y^c}{\partial A^k_{ij}}
\]
where $y^c$ is model prediction and $A^k$ is $k$-th channel of features from last layer. As IQA model has only one output, $c$ equals 1. To ensure differentiability, we detached $\alpha^c_l$ from the computational graph, calculating gradients exclusively over the channel maps. This modification is integral to the Baseline-GradCAM-Loss model.

We used 80\% of the KonIQ-10k dataset for training and the remaining 20\% for testing. Additionally, we evaluated our model on the CLIVE, SALICON, and CAT2000 datasets. Since KonIQ-10k does not provide saliency maps, we generated proxy SM using the MSINet model. Each experiment was repeated 10 times to ensure consistency, with the averaged results presented in Table 1.

We compared the dual-task models with the baseline model trained solely on IQA using the saliency extraction technique, Center Prior, and other State-of-the-Art saliency models. Our findings reveal that both dual-task approaches allow for effective training on IQA and SP tasks concurrently. These approaches maintain the performance of IQA models while significantly improving the results for SP.



\subsection{Non-uniform-Compression Dataset}

\begin{table*}[tbp]
\tabcolsep=5pt
\caption{
Simultaneous learning of two tasks.
}
\centering
\begin{tabular}{lcccccccccccc}
\toprule
\multirow{4}{*}{Method} & \multicolumn{4}{c}{IQA}                                   & \multicolumn{8}{c}{Saliency Prediction}                   \\ \cmidrule(lr){2-5} \cmidrule(lr){6-13}
                         & \multicolumn{2}{c}{KonIQ-10k\cite{koniq10k}} & \multicolumn{2}{c}{CLIVE\cite{clive}} & \multicolumn{4}{c}{SALICON\cite{salicon}} & \multicolumn{4}{c}{CAT2000\cite{cat2000}} \\ \cmidrule(lr){2-3} \cmidrule(lr){4-5} \cmidrule(lr){6-9} \cmidrule(lr){10-13}
  & SROCC $\uparrow$ & PLCC $\uparrow$ & SROCC $\uparrow$ & PLCC $\uparrow$ & NSS $\uparrow$ & SIM $\uparrow$ & CC $\uparrow$ & KLD $\downarrow$ & NSS $\uparrow$ & SIM $\uparrow$ & CC $\uparrow$ & KLD $\downarrow$ \\ \midrule
                      
Center Prior          & -              & -            & -            & -           & 0.578  & 0.536 & 0.544 & 0.779  & 0.303  & 0.599 & 0.771 & 0.672  \\ \cmidrule(lr){1-1} \cmidrule(lr){2-3} \cmidrule(lr){4-5} \cmidrule(lr){6-9} \cmidrule(lr){10-13}
Baseline              & \underline{0.913}           & \underline{0.931}         & \textbf{0.862}         & \textbf{0.848}    & 0.604  & 0.584 & 0.637 & 0.700   & 0.308  & 0.613 & 0.781 & 0.585  \\
Baseline-Sal-Loss     & \textbf{0.914}           & \textbf{0.930}        & \textit{0.845}         & \textit{0.839}    & \textbf{0.618}  & 0.688 & 0.803 & 0.429  & \underline{0.372}  & \textit{0.665} & 0.818 & 0.433  \\
Baseline-GradCAM-Loss & \textit{0.912}           & \underline{0.931}         & \underline{0.852}         & \underline{0.840}    & \underline{0.615}  & \textit{0.703} & \textit{0.825} & \textit{0.396}  & \textbf{0.382}  & \textbf{0.676} & \underline{0.826} & \textbf{0.403}  \\ \cmidrule(lr){1-1} \cmidrule(lr){2-3} \cmidrule(lr){4-5} \cmidrule(lr){6-9} \cmidrule(lr){10-13}
MSINet\cite{msinet}                & -              & -            & -            & -           & 0.612  & \underline{0.767} & \underline{0.891} & \textbf{0.252}  & \textit{0.370}  & 0.664 & \textit{0.820} & \textit{0.421}  \\
TranSalNet\cite{transalnet}            & -              & -            & -            & -           & \textit{0.613}  & \textbf{0.776} & \textbf{0.903} & \underline{0.258}  & 0.369  & \underline{0.670} & \textbf{0.829} & \underline{0.407}  \\
\bottomrule
\end{tabular}
\label{tab:multitask}
\end{table*}



We hypothesized that the existing IQA datasets lack sufficient complexity (due to almost-uniform compression and degradation), hindering the ability to detect emerging capabilities in IQA models, utilizing saliency. Consequently, we constructed a new dataset of high-quality images compressed with nonuniform codecs called Saliency-Aware Compressed Images Dataset (SACID). Specifically, we employed the custom codec~\cite{salx264} and sourced 50 images from the CLIC-2021 dataset. 

We generated saliency maps for each image using the TranSalNet model and applied compression with four presets: one without saliency consideration and three that incorporate saliency. For each preset, we selected three bitrates, yielding bit-per-pixel (bpp) values of 0.08, 0.12, and 0.16 -- similar to those in the CLIC-2021 challenge.
Our saliency-aware compression used settings of \((\verb|saliency_s0|, \verb|saliency_bitrate|) \in \left[(75, 80), (60, 80), (60, 40) \right]\), creating a diverse set of images. We generated a total of 720 nonuniformly compressed images for evaluation. 

To obtain subjective scores, we conducted pairwise comparisons using over 1,400 assessors from the crowdsourcing platform \href{https://www.subjectify.us}{\texttt{Subjectify.us}}. Assessors were presented with 3 verification and 25 random pairs of images with different picture qualities and were asked the following question: ``You will be shown sequential pairs of images with different picture quality. For each pair, select the image that has the most acceptable quality for viewing, or note that the quality in this pair is almost the same''.

We evaluated all models from the PYIQA toolbox~\cite{pyiqa}, saliency-aware versions of PSNR and SSIM (EW-PSNR and EW-SSIM~\cite{ew_ssim}), our model variants (Baseline, Baseline-Sal-Loss, and Baseline-GradCAM-Loss), and salient deep-learning models (SGDNet\cite{sgdnet}, etc.). Models were compared in terms of SROCC, PLCC and fraction of concordant pairs (FracCP). To calculate FracCP, groups corresponding to different images were considered, and in each group, the fraction of ordered pairs was counted. The resulting numbers were averaged across all groups.

Results are listed in Tab.~\ref{tab:nonuni_res}. Notably, salient versions of PSNR and SSIM remarkably outperformed the originals. But Baseline, Baseline-Sal-Loss, and Baseline-GradCAM-Loss models demonstrated similar performance, implying that saliency fails to enhance deep-learning metrics significantly. We attribute this to the reliance of methods on conventional datasets, where most distortions are uniform, despite being trained with saliency. Thus, IQA models are limited by current compression standards and may show lower correlations in the non-uniform compression domain.


\section{Discussion}


\begin{table}[tbp]
\tabcolsep=5pt
\centering
\caption{Quantitative results on SACID.
}
\begin{tabular}{clccc}
\toprule
\multicolumn{1}{c}{Type}                       & Method                 & SROCC $\uparrow$ & PLCC $\uparrow$ & FracCP $\uparrow$ \\ \midrule
\multicolumn{1}{l}{\multirow{7}{*}{\thead{NSS \\ based}}} 
                                               & MS-SSIM~\cite{ms_ssim} & 0.807  & 0.849 & 0.507 \\
\multicolumn{1}{l}{}                           & BRISQUE~\cite{brisque} & 0.817  & 0.833 & \textit{0.572} \\
\multicolumn{1}{l}{}                           & PSNR                   & 0.822  & 0.858 & 0.505 \\
\multicolumn{1}{l}{}                           & SSIM~\cite{ssim}       & 0.835  & \textit{0.880} & 0.502 \\
\multicolumn{1}{l}{}                           & VIF~\cite{vif}         & \textit{0.848}  & \underline{0.891} & 0.526 \\
\multicolumn{1}{l}{}                           & EW-SSIM~\cite{ew_ssim} & \underline{0.850}  & 0.867 & \underline{0.636} \\
\multicolumn{1}{l}{}                           & EW-PSNR~\cite{ew_ssim} & \textbf{0.875}  & \textbf{0.893} & \textbf{0.678} \\ \cmidrule(lr){1-5}
\multirow{5}{*}{FR}                            
                                               & AHIQ~\cite{ahiq}       & 0.789  & 0.817 & \textit{0.546} \\
                                               & LPIPS~\cite{lpips}     & 0.818  & \textit{0.854} & 0.512 \\
                                               & PieAPP~\cite{pieapp}   & \textit{0.833 } & \underline{0.869} & 0.535 \\
                                               & TOPIQ-FR~\cite{topiq}  & \underline{0.836}  & 0.848 & \textbf{0.629} \\
                                               & DISTS~\cite{dists}     & \textbf{0.871}  & \textbf{0.904} & \underline{0.589} \\ \cmidrule(lr){1-5}
\multirow{11}{*}{NR}                           
                                               & MANIQA~\cite{maniqa}   & 0.752  & 0.773 & 0.564 \\
                                               & TReS~\cite{tres}       & 0.765  & 0.800 & 0.559 \\
                                             & HyperIQA~\cite{hyperiqa} & 0.784  & 0.802 & 0.562 \\
                                            & SGDNet-None~\cite{sgdnet} & 0.820  & 0.853 & 0.651 \\
                                               & PaQ2PiQ~\cite{paq2piq} & 0.848  & 0.858 & 0.619 \\
                                          & SGDNet-Output~\cite{sgdnet} & 0.854  & 0.879 & \textbf{0.687} \\
                                               & Baseline-GradCAM-Loss  & 0.863  & 0.865 & 0.624 \\
                                             & CLIP-IQA+~\cite{clipiqa} & 0.869 & \textit{0.896} & 0.598 \\
                                               & Baseline               & 0.871  & 0.881 & 0.609 \\
                                               & TOPIQ-NR~\cite{topiq}  & 0.876  & 0.886 & 0.629 \\
                                               & Baseline-Sal-Loss      & \textit{0.880}  & 0.886 & 0.630 \\
                                               & MUSIQ~\cite{musiq}     & \underline{0.888}  & \underline{0.899} & \textit{0.662} \\
                                               & DBCNN~\cite{dbcnn}     & \textbf{0.901}  & \textbf{0.917} & \underline{0.666} \\
\bottomrule
\end{tabular}
\label{tab:nonuni_res}
\end{table}

Our experiments reveal a clear link between saliency prediction and IQA. Using GradCAM to extract saliency maps from IQA models shows they implicitly capture human attention better than simple baselines like center-prior. Our dual-task training approach demonstrates that explicitly using saliency maintains or slightly improves IQA performance without adding model complexity.

Masking experiments confirm the significance of salient image regions, as masking them significantly reduces IQA scores. However, directly incorporating saliency into neural IQA methods showed limited gains on our non-uniform compression dataset (SACID). This limitation likely stems from existing IQA datasets being dominated by uniform distortions, restricting models' ability to leverage saliency cues effectively. 
Therefore, building datasets with diverse, saliency-driven, and non-uniform distortions is important for further research.


\section{Conclusion}

This study examined the relationship between saliency prediction and image-quality assessment. We propose a technique for the extraction of saliency maps from IQA models and empirically demonstrate their reliance on salient regions. Additionally, we propose a parameter-free dual-task learning approach without sacrificing quality. Finally, we curated a dataset of non-uniformly compressed images and performed a large-scale comparison. We conclude that saliency can remarkably improve NSS-based methods, while learning-based methods face slight improvement since they already take saliency priors into account during conventional training.

\section*{Acknowledgments} This research was supported by Russian Science Foundation under grant 24-21-00172, \href{https://rscf.ru/en/project/24-21-00172/}{https://rscf.ru/en/project/24-21-00172/} and was carried out using the MSU-270 supercomputer of Lomonosov Moscow State University.


\newpage

\bibliographystyle{IEEEtran}
\bibliography{bibliography}

\begin{thebibliography}{10}
\providecommand{\url}[1]{#1}
\csname url@samestyle\endcsname
\providecommand{\newblock}{\relax}
\providecommand{\bibinfo}[2]{#2}
\providecommand{\BIBentrySTDinterwordspacing}{\spaceskip=0pt\relax}
\providecommand{\BIBentryALTinterwordstretchfactor}{4}
\providecommand{\BIBentryALTinterwordspacing}{\spaceskip=\fontdimen2\font plus
\BIBentryALTinterwordstretchfactor\fontdimen3\font minus \fontdimen4\font\relax}
\providecommand{\BIBforeignlanguage}[2]{{%
\expandafter\ifx\csname l@#1\endcsname\relax
\typeout{** WARNING: IEEEtran.bst: No hyphenation pattern has been}%
\typeout{** loaded for the language `#1'. Using the pattern for}%
\typeout{** the default language instead.}%
\else
\language=\csname l@#1\endcsname
\fi
#2}}
\providecommand{\BIBdecl}{\relax}
\BIBdecl

\bibitem{saliency_bio}
R.~Desimone and J.~Duncan, ``Neural mechanisms of selective visual attention,'' \emph{Annual review of neuroscience}, 1995.

\bibitem{iqa_hvs}
X.~Gao, W.~Lu, D.~Tao, and X.~Li, ``Image quality assessment and human visual system,'' in \emph{Visual Communications and Image Processing 2010}, 2010.

\bibitem{transalnet}
J.~Lou, H.~Lin, D.~Marshall, D.~Saupe, and H.~Liu, ``{TranSalNet}: {Towards} perceptually relevant visual saliency prediction,'' \emph{Neurocomputing}, 2022.

\bibitem{survey_iqa}
L.~Wang, ``A survey on iqa,'' \emph{arXiv preprint arXiv:2109.00347}, 2021.

\bibitem{ssim}
Z.~Wang, A.~Bovik, H.~Sheikh, and E.~Simoncelli, ``Image quality assessment: from error visibility to structural similarity,'' \emph{IEEE Transactions on Image Processing}, 2004.

\bibitem{ms_ssim}
Z.~Wang, E.~P. Simoncelli, and A.~C. Bovik, ``Multiscale structural similarity for image quality assessment,'' in \emph{The Thrity-Seventh Asilomar Conference on Signals, Systems \& Computers, 2003}, 2003.

\bibitem{vif}
H.~R. Sheikh and A.~C. Bovik, ``A visual information fidelity approach to video quality assessment,'' in \emph{The first international workshop on video processing and quality metrics for consumer electronics}, 2005.

\bibitem{brisque}
A.~Mittal, A.~K. Moorthy, and A.~C. Bovik, ``No-reference image quality assessment in the spatial domain,'' \emph{IEEE Transactions on image processing}, 2012.

\bibitem{ew_ssim}
H.~Liu and I.~Heynderickx, ``\BIBforeignlanguage{en}{Studying the added value of visual attention in objective image quality metrics based on eye movement data},'' in \emph{\BIBforeignlanguage{en}{2009 16th {IEEE} {International} {Conference} on {Image} {Processing} ({ICIP})}}, 2009.

\bibitem{hyperiqa}
S.~Su, Q.~Yan, Y.~Zhu, C.~Zhang, X.~Ge, J.~Sun, and Y.~Zhang, ``\BIBforeignlanguage{en}{Blindly assess image quality in the wild guided by a self-adaptive hyper network},'' in \emph{\BIBforeignlanguage{en}{Proceedings of the IEEE/CVF conference on computer vision and pattern recognition}}, 2020.

\bibitem{musiq}
J.~Ke, Q.~Wang, Y.~Wang, P.~Milanfar, and F.~Yang, ``\BIBforeignlanguage{en}{{MUSIQ}: {Multi}-scale {Image} {Quality} {Transformer}},'' in \emph{\BIBforeignlanguage{en}{2021 {IEEE}/{CVF} {International} {Conference} on {Computer} {Vision} ({ICCV})}}, 2021.

\bibitem{topiq}
C.~Chen, J.~Mo, J.~Hou, H.~Wu, L.~Liao, W.~Sun, Q.~Yan, and W.~Lin, ``{TOPIQ}: {A} {Top}-down {Approach} from {Semantics} to {Distortions} for {Image} {Quality} {Assessment},'' 2023.

\bibitem{halfimage}
J.~You, Y.~Lin, and J.~Korhonen, ``Half of an image is enough for quality assessment,'' 2023.

\bibitem{wang_ugc_2023}
X.~Wang, A.~Katsenou, and D.~Bull, ``{UGC} {Quality} {Assessment}: {Exploring} the {Impact} of {Saliency} in {Deep} {Feature}-{Based} {Quality} {Assessment},'' 2023.

\bibitem{iqa_sal}
H.~Liu, U.~Engelke, J.~Wang, P.~Le~Callet, and I.~Heynderickx, ``How does image content affect the added value of visual attention in objective image quality assessment?'' \emph{IEEE Signal Processing Letters}, 2013.

\bibitem{hantao_liu_visual_2011}
{Hantao Liu} and I.~Heynderickx, ``\BIBforeignlanguage{en}{Visual {Attention} in {Objective} {Image} {Quality} {Assessment}: {Based} on {Eye}-{Tracking} {Data}},'' \emph{\BIBforeignlanguage{en}{IEEE Transactions on Circuits and Systems for Video Technology}}, 2011.

\bibitem{lin_saliency-aware_2023}
L.~Lin, Y.~Zheng, W.~Chen, C.~Lan, and T.~Zhao, ``Saliency-{Aware} {Spatio}-{Temporal} {Artifact} {Detection} for {Compressed} {Video} {Quality} {Assessment},'' 2023.

\bibitem{sgdnet}
S.~Yang, Q.~Jiang, W.~Lin, and Y.~Wang, ``\BIBforeignlanguage{en}{{SGDNet}: {An} {End}-to-{End} {Saliency}-{Guided} {Deep} {Neural} {Network} for {No}-{Reference} {Image} {Quality} {Assessment}},'' in \emph{\BIBforeignlanguage{en}{Proceedings of the 27th {ACM} {International} {Conference} on {Multimedia}}}, 2019.

\bibitem{hvs5m}
A.-X. Zhang, Y.-G. Wang, W.~Tang, L.~Li, and S.~Kwong, ``{HVS} {Revisited}: {A} {Comprehensive} {Video} {Quality} {Assessment} {Framework},'' 2022.

\bibitem{cai_blind_2023}
R.~Cai and M.~Fang, ``\BIBforeignlanguage{en}{Blind image quality assessment by simulating the visual cortex},'' \emph{\BIBforeignlanguage{en}{The Visual Computer}}, 2023.

\bibitem{sal_transformer}
M.~Zhu, G.~Hou, X.~Chen, J.~Xie, H.~Lu, and J.~Che, ``\BIBforeignlanguage{en}{Saliency-{Guided} {Transformer} {Network} combined with {Local} {Embedding} for {No}-{Reference} {Image} {Quality} {Assessment}},'' in \emph{\BIBforeignlanguage{en}{2021 {IEEE}/{CVF} {International} {Conference} on {Computer} {Vision} {Workshops} ({ICCVW})}}, 2021.

\bibitem{scvs}
J.~Ji, K.~Xiang, and X.~Wang, ``\BIBforeignlanguage{en}{{SCVS}: blind image quality assessment based on spatial correlation and visual saliency},'' \emph{\BIBforeignlanguage{en}{The Visual Computer}}, 2023.

\bibitem{lpips}
R.~Zhang, P.~Isola, A.~A. Efros, E.~Shechtman, and O.~Wang, ``The unreasonable effectiveness of deep features as a perceptual metric,'' in \emph{Proceedings of the IEEE Conference CVPR}, 2018.

\bibitem{cam}
B.~Zhou, A.~Khosla, A.~Lapedriza, A.~Oliva, and A.~Torralba, ``Learning {Deep} {Features} for {Discriminative} {Localization},'' 2015.

\bibitem{gradcam}
R.~R. Selvaraju, M.~Cogswell, A.~Das, R.~Vedantam, D.~Parikh, and D.~Batra, ``Grad-cam: Visual explanations from deep networks via gradient-based localization,'' in \emph{Proceedings of the IEEE international conference on computer vision}, 2017.

\bibitem{gradcam_pp}
A.~Chattopadhay, A.~Sarkar, P.~Howlader, and V.~N. Balasubramanian, ``Grad-cam++: Generalized gradient-based visual explanations for deep convolutional networks,'' in \emph{2018 IEEE winter conference on applications of computer vision (WACV)}, 2018.

\bibitem{hirescam}
R.~L. Draelos and L.~Carin, ``Use {HiResCAM} instead of {Grad}-{CAM} for faithful explanations of convolutional neural networks,'' 2021.

\bibitem{normgrad}
S.-A. Rebuffi, R.~Fong, X.~Ji, H.~Bilen, and A.~Vedaldi, ``{NormGrad}: {Finding} the {Pixels} that {Matter} for {Training},'' 2019.

\bibitem{sanity_checks_xai}
R.~Tomsett, D.~Harborne, S.~Chakraborty, P.~Gurram, and A.~Preece, ``Sanity {Checks} for {Saliency} {Metrics},'' 2019.

\bibitem{salicon}
M.~Jiang, S.~Huang, J.~Duan, and Q.~Zhao, ``Salicon: Saliency in context,'' in \emph{The IEEE Conference on Computer Vision and Pattern Recognition (CVPR)}, 2015.

\bibitem{dbcnn}
W.~Zhang, K.~Ma, J.~Yan, D.~Deng, and Z.~Wang, ``Blind image quality assessment using a deep bilinear convolutional neural network,'' \emph{IEEE Transactions on Circuits and Systems for Video Technology}, 2018.

\bibitem{clipiqa}
J.~Wang, K.~C.~K. Chan, and C.~C. Loy, ``Exploring {CLIP} for {Assessing} the {Look} and {Feel} of {Images},'' in \emph{Proceedings of the AAAI Conference on Artificial Intelligence}, 2022.

\bibitem{msinet}
A.~Kroner, M.~Senden, K.~Driessens, and R.~Goebel, ``Contextual {Encoder}-{Decoder} {Network} for {Visual} {Saliency} {Prediction},'' \emph{Neural Networks}, 2020.

\bibitem{efficientnet}
M.~Tan and Q.~Le, ``Efficientnet: Rethinking model scaling for convolutional neural networks,'' in \emph{International conference on machine learning}, 2019.

\bibitem{resnet}
K.~He, X.~Zhang, S.~Ren, and J.~Sun, ``Deep residual learning for image recognition,'' in \emph{Proceedings of the IEEE conference on computer vision and pattern recognition}, 2016.

\bibitem{jacobgilpytorchcam}
J.~Gildenblat and contributors, ``Pytorch library for cam methods,'' \url{https://github.com/jacobgil/pytorch-grad-cam}, 2021.

\bibitem{lyudvichenko2017semiautomatic}
V.~Lyudvichenko, M.~Erofeev, Y.~Gitman, and D.~Vatolin, ``A semiautomatic saliency model and its application to video compression,'' in \emph{2017 13th IEEE International Conference on Intelligent Computer Communication and Processing (ICCP)}.\hskip 1em plus 0.5em minus 0.4em\relax IEEE, 2017, pp. 403--410.

\bibitem{koniq10k}
V.~{Hosu}, H.~{Lin}, T.~{Sziranyi}, and D.~{Saupe}, ``Koniq-10k: An ecologically valid database for deep learning of blind image quality assessment,'' \emph{IEEE Transactions on Image Processing}, 2020.

\bibitem{clive}
D.~Ghadiyaram and A.~C. Bovik, ``Massive online crowdsourced study of subjective and objective picture quality,'' \emph{IEEE Transactions on Image Processing}, 2015.

\bibitem{cat2000}
A.~Borji and L.~Itti, ``Cat2000: A large scale fixation dataset for boosting saliency research. arxiv 2015,'' \emph{arXiv preprint arXiv:1505.03581}, 2019.

\bibitem{salx264}
V.~Lyudvichenko, M.~Erofeev, Y.~Gitman, and D.~Vatolin, ``A semiautomatic saliency model and its application to video compression,'' in \emph{2017 13th IEEE International Conference on Intelligent Computer Communication and Processing (ICCP)}, 2017.

\bibitem{pyiqa}
C.~Chen, ``{IQA PyTorch},'' \url{https://github.com/chaofengc/IQA-PyTorch}, 2021.

\bibitem{ahiq}
S.~Lao, Y.~Gong, S.~Shi, S.~Yang, T.~Wu, J.~Wang, W.~Xia, and Y.~Yang, ``Attentions help cnns see better: Attention-based hybrid image quality assessment network,'' in \emph{Proceedings of the IEEE/CVF conference on computer vision and pattern recognition}, 2022.

\bibitem{pieapp}
E.~Prashnani, H.~Cai, Y.~Mostofi, and P.~Sen, ``Pieapp: Perceptual image-error assessment through pairwise preference,'' in \emph{Proceedings of the IEEE Conference on Computer Vision and Pattern Recognition}, 2018.

\bibitem{dists}
K.~Ding, K.~Ma, S.~Wang, and E.~P. Simoncelli, ``Image quality assessment: Unifying structure and texture similarity,'' \emph{arXiv:2004.07728}, 2020.

\bibitem{maniqa}
S.~Yang, T.~Wu, S.~Shi, S.~Lao, Y.~Gong, M.~Cao, J.~Wang, and Y.~Yang, ``{MANIQA}: {Multi}-dimension {Attention} {Network} for {No}-{Reference} {Image} {Quality} {Assessment},'' 2022.

\bibitem{tres}
S.~A. Golestaneh, S.~Dadsetan, and K.~M. Kitani, ``No-reference image quality assessment via transformers, relative ranking, and self-consistency,'' in \emph{Proceedings of the IEEE/CVF winter conference on applications of computer vision}, 2022.

\bibitem{paq2piq}
Z.~Ying, H.~Niu, P.~Gupta, D.~Mahajan, D.~Ghadiyaram, and A.~C. Bovik, ``From patches to pictures (paq-2-piq): Mapping the perceptual space of picture quality,'' in \emph{2020 {IEEE/CVF} Conference {CVPR}}, 2020.

\end{thebibliography}

\end{document}